\documentclass[10pt,twocolumn,a4paper]{article}

\pdfoutput=1

\usepackage{cvpr}
\usepackage{times}
\usepackage{epsfig}
\usepackage{graphicx}
\usepackage{amsmath}
\usepackage{amssymb}

\usepackage[pagebackref=true,breaklinks=true,letterpaper=true,colorlinks,bookmarks=false]{hyperref}

 \cvprfinalcopy 

\def\cvprPaperID{6179} 

\newcommand{\lambdaF}{\lambda_F}
\newcommand{\Fsched}{F_\text{sched}}
\newcommand{\rhomax}{\rho_\text{max}}
\newcommand{\mlr}{\mu}
\newcommand{\clip}{\text{clip}}

\newcommand{\eps}{\epsilon}

\newcommand{\dd}{\partial}
\newcommand{\simiid}{\overset{\text{i.i.d.}}\sim}
\newcommand{\ve}[1]{\boldsymbol{#1}}
\newcommand{\abs}[1]{\lvert{#1}\rvert}
\newcommand{\textem}[1]{\textit{#1}}

\newcommand{\EE}{\mathop{\mathrm{E}}}

\begin{document}

\title{Channel-wise pruning of neural networks with tapering resource constraint}

\author{Alexey Kruglov\\
Intel Corp.\\
{\tt\small alexey.kruglov@gmail.com}
}

\maketitle

\begin{abstract}
Neural network pruning is an important step in design process of efficient neural
networks for edge devices with limited computational power.
Pruning is a form of knowledge transfer from the weights of the original network
to a smaller target subnetwork.

We propose a new method for compute-constrained structured channel-wise pruning of convolutional neural networks.
The method iteratively fine-tunes the network, while gradually tapering the computation resources available to the pruned network
via a holonomic constraint in the method of Lagrangian multipliers framework.
An explicit and adaptive automatic control over the rate of tapering is provided.
The trainable parameters of our pruning method are separate from the weights of the neural network,
which allows us to avoid the interference with the neural network solver
(e.g. avoid the direct dependence of pruning speed on neural network learning rates).

Our method combines the ``rigoristic'' approach by the direct application of constrained optimization,
avoiding the pitfalls of ADMM-based methods, like their need to define the target amount of resources
for each pruning run, and direct dependence of pruning speed and priority of pruning on the relative scale
of weights between layers.

For VGG-16~@ ILSVRC-2012, we achieve reduction of $15.47 \to 3.87$~GMAC with only 1\% top-1 accuracy
reduction ($68.4\% \to 67.4\%$).
For AlexNet~@ ILSVRC-2012, we achieve $0.724\to0.411$~GMAC with 1\% top-1 accuracy reduction ($56.8\% \to 55.8\%$).
\end{abstract}

\section{Introduction}

Convolutional neural networks (CNNs) became a practical tool for computer vision applications.
This drives efforts for optimization of neural networks either due to hardware-limited computational power
for edge devices, or for economical reasons for server segment.

CNNs are known to demonstrate quality-performance tradeoff: facing limited computation resources 
one is generally limited to network topologies providing lower quality.~\cite{bib:mobilenet}
The ways around this are: code optimization, network quantization, methods employing low-rank matrix decompositions
and neural network pruning.
We focus on pruning, which is a repeated removal of a subset of network elements 
(weights, kernels or channels, depending on granularity level~\cite{bib:Mao2017})
and fine-tuning.
Pruning with weight or kernel granularities results in networks with sparse weight matrices,
which require effective implementation of sparse convolutional layer on target hardware platforms.
Weight granularity pruning achieves smaller network FLOPs for the same quality compared to channel granularity,~\cite{bib:Mao2017}
because they can leverage sub-channel sparsity.
We focus on pruning with channel granularity (channel-wise pruning), which results in dense weight matrices in the pruned networks,
because this doesn't require the costly reimplementation of convolutional layers.

From the \textit{problem statement} point, the word ``pruning'' may refer to rather different tasks.
They differ in a number of ``dimensions'', among them:
\begin{itemize}
\item
Some papers focus on the number of parameters after pruning to optimize the model size or memory bandwidth, 
while the other focus on the number of floating-point operations (FLOPs) or inference time.
The difference is especially significant for classification networks with large fully connected (FC) layers (AlexNet, VGG),
with most parameters in FC layers and most FLOPs in convolutional layers.

\item
Some methods leave decision about the distribution of pruned elements between layers outside their scope.
This either requires manual trial-and-error experimentation, or an external automated method for this.

\item
Another ``dimension'' is the way the amount of pruning is controlled:
some methods provide FLOPs-quality curve after a single run (mostly iterative methods based on~\cite{bib:Han2015}), 
other methods require the target number of FLOPs as a method input,
or even take a parameter controlling final FLOPs in a complex way (e.g. \cite{bib:Ye2018}).
Since each run usually includes heavy fine-tuning, this may be important, 
\end{itemize}

\begin{figure*}[t]
\begin{center}
\includegraphics[width=12cm]{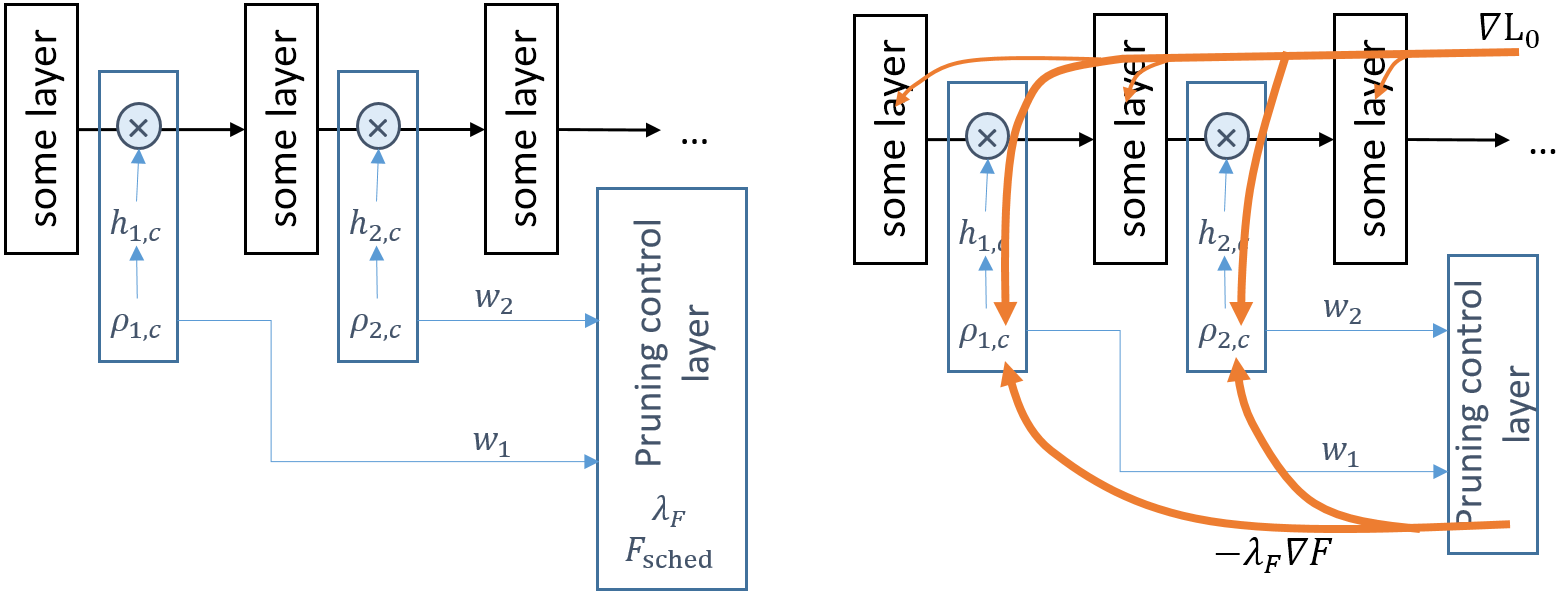}
\end{center}
   \caption{Block diagram of a neural network instrumented with pruning layers. 
   Left: network layers and forward pass data flow, right: backpropagation data flow is shown in orange. 
   Black boxes are the layers of the network being pruned, blue boxes are pruning layers inserted at the 
   desired pruning sites, and a single Pruning control layer, that is responsible 
   for updating $\lambdaF$ and $\Fsched$, and backpropagating the $-\lambdaF \nabla F$ term of loss gradient.}
\label{fig:block-diag}
\end{figure*}
Relative to this framework,
our method is designed to control FLOPs consumption by the inference of the pruned network,
to distribute the channels between layers during pruning,
and to provide FLOPs--quality curve after a single run.

\medskip
We consider resource-aware pruning as a constrained optimization problem with a moving holonomic constraint:
we optimize both neural network weights $\ve\theta$ and the parameters $\ve\rho$ defining pruning probabilities $1-\ve p(\ve\rho)$,
while keeping the estimated FLOPs consumption $F(\ve p)$ by the pruned network close to the schedule~$\Fsched(i)$:
\begin{equation}  \label{eq:constr}
F(\ve p^i) \approx \Fsched(i)  \text,
\end{equation}
where $i$ is iteration number, and $\ve p^i$ is $\ve p$ for the $i$-th iteration.
We convert this constrained optimization problem to an unconstrained one with the method of Lagrange multipliers
by adding a Lagrangian term to the loss function~$L_0$:
\begin{equation}  \label{eq:lagr}
L(\ve\theta, \ve\rho) = L_0(\ve\theta, \ve\rho) - \lambdaF F(\ve \rho)  \text.
\end{equation}
and updating $\lambdaF$ depending on $(F(\ve p^i)-\Fsched(i))$ to keep~\eqref{eq:constr}, details below.

Since slowdown of schedule $\Fsched(i)$ at the later iterations of pruning defers the onset of quality degradation, 
but the values of FLOPs~$F$ or iterations~$i$ where this happens are not known in advance,
our method controls $\Fsched(i)$ by feedback from $\lambdaF$ based on
\begin{equation}  \label{eq:Fsched0}
\Fsched(i+1) = \Fsched(i) - \frac{\mlr}{\abs{\lambdaF}}
\end{equation}
with some \textbf{modifications,} where $\mlr$ is a hyperparameter. 
The intuition is that $\abs{\lambdaF}$ grows when quality(FLOPs) curve starts to fall quicker,
so we allocate more fine-tuning time there.


\subsection{Related works}



One large branch of pruning methods stems from the basic scheme of Han et~al. (2015)~\cite{bib:Han2015},
we'll call them \textem{``heuristic methods''.}
These methods repeatedly choose elements
based on some scalar metric (salience), and remove them from the network.
Each iteration of removal is followed by fine-tuning.
Salience can be based on 
  $\ell_1$~norm of element weights \cite{bib:Han2015, bib:Srinivas2015, bib:Li2016, bib:ZWang2017GGP, bib:Guo2016, bib:Yang2016},
  Taylor estimate of change in loss from element removal \cite{bib:Molchanov2016},
  percentage of zeros in channel weights (APoZ metric~\cite{bib:Hu2016}),
  statistics of channels activations \cite{bib:Polyak2015},
  etc.
Some methods improve fine-tuning by compensating removal of elements through changing the remaining weights in the network:
  by using linear least squares to approximate the output of the original layer in $\ell_2$ metric~\cite{bib:Yang2016, bib:Polyak2015}; 
  or finding paired channels with similar weights and updating weights of one channel to compensate for the removal of the other~\cite{bib:Srinivas2015}.
Another way to help fine-tuning is by making pruning reversible (``splicing''): \cite{bib:Guo2016, bib:He2018}.

To introduce ``awareness'' of FLOPs or other resources, some methods inject a resource-dependent term 
into salience~\cite{bib:Molchanov2016, bib:Yang2016, bib:Wolfe2017},
or by more complex means~\cite{bib:Park2016}.

\medskip
``Fisher pruning''~\cite{bib:Theis2018} resembles these ``heuristic'' methods,
but its salience is based on the method of Lagrange multipliers,
which makes this method resource-aware and less heuristic.
It removes a channel every pruning iteration, so its pruning speed is fixed and doesn't slow down.

\medskip
Another group of methods is based on \textem{constrained optimization} 
with Alternating Direction Method of Multipliers (ADMM).

Carreira-Perpi\~n\'an et~al. (2017, 2018) introduced an ADMM-based learning-pruning method~\cite{bib:Carreira2017a, bib:Carreira2018}, 
where a model is trained from random initialized weights to a constrained number of non-zero weights.
Zhang et al. (2018) introduces a similar ADMM-based pruning method~\cite{bib:Zhang2018a, bib:Zhang2018b}.
Both are weight granularity methods. 
Both methods add a term to the loss function, that draws weights to their projection onto the subset with a limited number of non-zero values.
They need the target number of pruned weights as a hyperparameter.
These methods can distribute pruned weights across layers by projecting in a joint space of the weights of all layers,
however this makes the priority of weight allocation between layers dependent on the relative scale of parameters
and on weight learning rates.
Beside this, these methods don't control the speed of pruning explicitly, which can result in 1) dependence of 
pruning speed on the overall weight scale, and 2) disbalance of pruning speeds between layers. 
Available weight solvers are limited, because they must be compatible with the pruning step.
These problems are caused by the dual role of weights in these methods: they define both the state of network training,
and the state of pruning.

\medskip
A group of optimization-based methods train channel multipliers with
and integer optimization methods (ISTA, heuristic, or ``reparametrization trick'' in our case).
They separate pruning parameters from network weights.

\begin{figure*}[t]
\begin{center}
\includegraphics[scale=0.65]{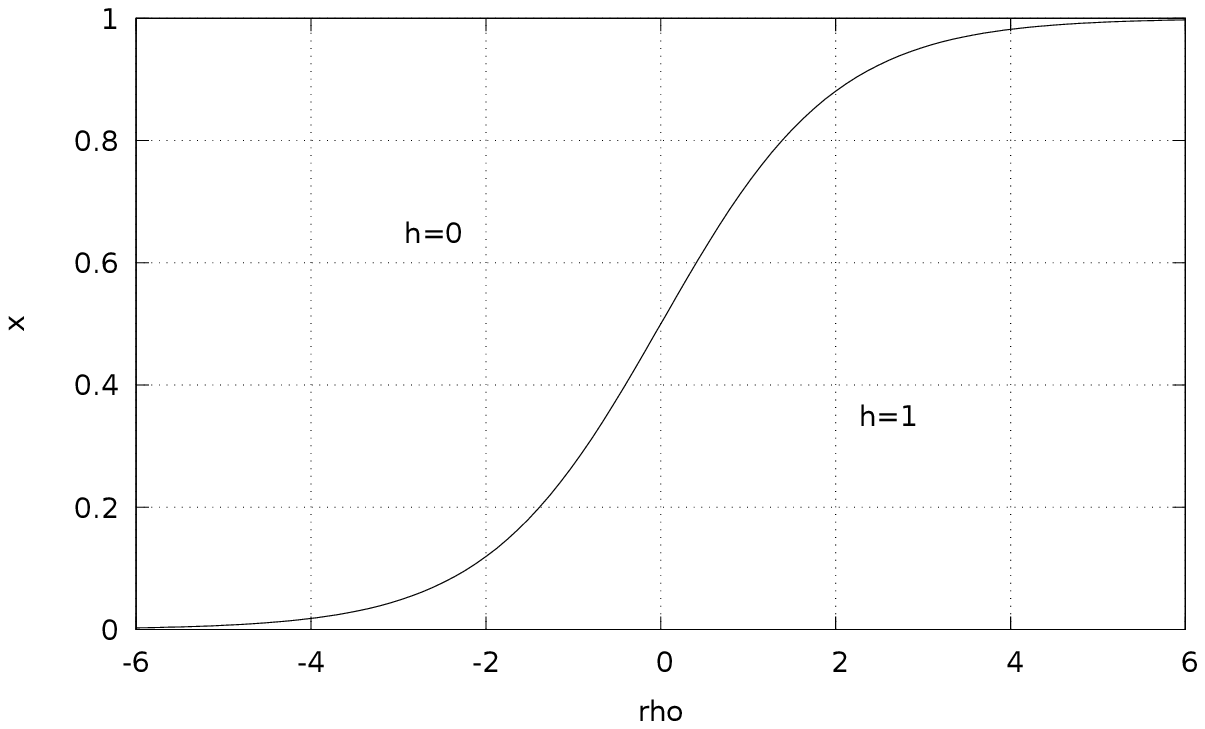}
\includegraphics[scale=0.65]{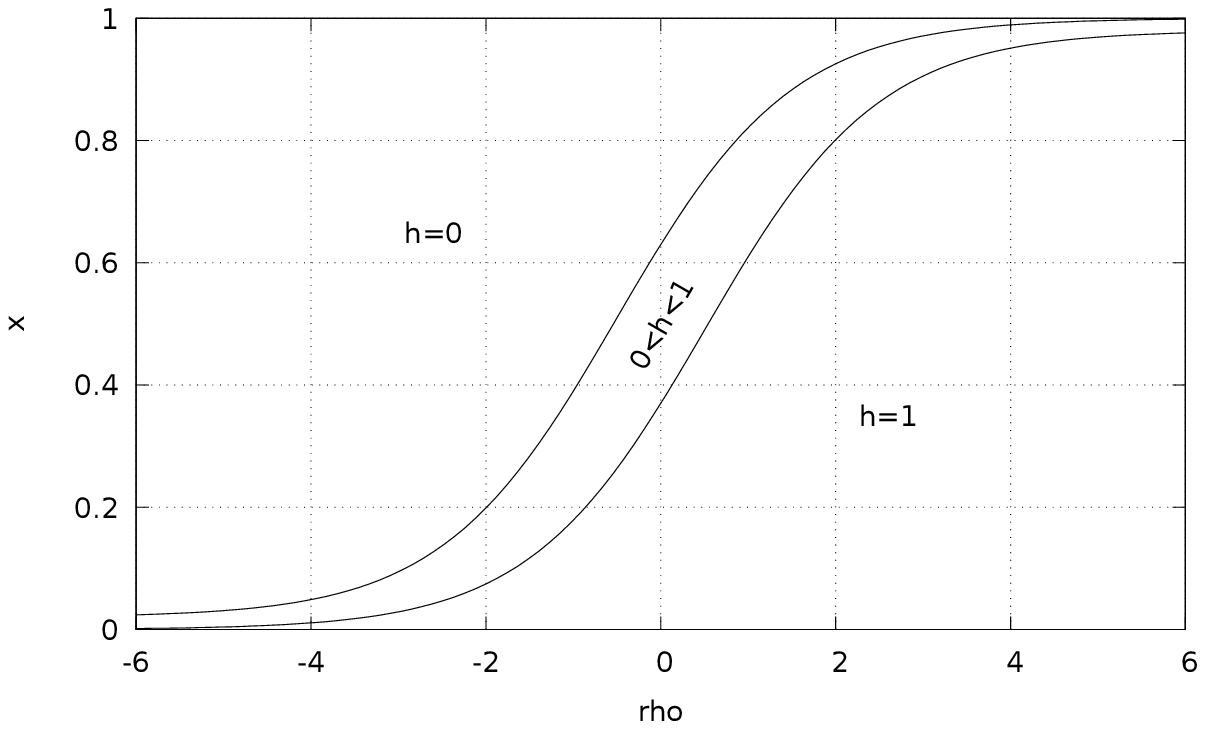}
\end{center}
   \caption{Graph showing $h(\rho,x)=0$ and $h(\rho,x)=1$ regions for $\eps=0$ (left) and for $\eps=0.5$ (right).}
\label{fig:rho-x}
\end{figure*}
%
The method from~\cite{bib:Ye2018} trains channel scaling factors to simulate channel granularity pruning, 
however the factors are not limited to $[0,1]$ range.
The factors are updated with an SGD-like method called ISTA, that includes a sparsity-inducing $\ell_1$ regularization
term resembling the Lagrangian term, which also makes this method resource-aware.


Structured Probabilistic Pruning~\cite{bib:HWang2017SPP} trains probabilities of channel removal.
Every pruning iteration the probabilities are updated with a heuristic rule based on the rank of the channel 
across all layers by $\ell_1$ metric of channel weights.
This requires the user to define the desired number of channels in advance. 
The method can provide size-quality curve based on the intermediate iterations and is not resource-aware.

Our method also trains per-channel parameters that define probabilities, but we use ``reparametrization trick'' to propagate
loss gradient to probabilities with backpropagation algorithm.


\section{Method}
We consider pruning as a constrained optimization problem with slowly tapering 
amount of available global resource $\Fsched(i)$, see~\eqref{eq:constr}.
Optimization starts from the pretrained weights $\ve\theta^{i=0} = \ve\theta_0$
and with all channels in place: $p^{i=0}=\sigma(\rho^{i=0})\approx1$, $\rho^{i=0}=\rhomax$.
%
Pruning parameters are updated every fine-tuning iteration.

\subsection{Notation}
\noindent
$i \geq 0$, iteration number \\
$n \in [1, N]$, sample index in a minibatch \\
$l$, pruning site index  \\
$c \in [1, n_l]$, index of channel in a pruning site \\
$p_{l,c}^i$, probability of retaining (not pruning) a channel \\
$\rho_{l,c}^i$, parameters defining $p_{l,c}^i = \sigma(\rho_{l,c}^i)$ \\
$\sigma(t) = 1/(1+e^{-t})$, sigmoid function \\
$x_{l,c}^{i,n} \simiid U(0,1)$, uniform random numbers in $[0,1]$ \\
$h_{l,c}^{i,n}$, channel scaling factors defined as 
\begin{equation}  \label{eq:h-def}
  h_{l,c}^{i,n} = h(\rho_{l,c}^i, x_{l,c}^{i,n}, \eps)
\end{equation} \\
$\eps$, a parameter: in $\eps\to0$ limit be get 
\begin{equation}  \label{eq:h-lim}
  h_{l,c}^{i,n} \simiid \text{Bernoulli}(p_{l,c}^i) \text, 
\end{equation}
i.e. scaling factors become 0 or~1 \\
$\ve\theta^i$, neural network weights \\
$X^{i,n},Y^{i,n}$, minibatch of samples from the dataset, $X$ for network inputs, $Y$ for ground truth data \\
$L(\ve\theta,\ve h,X,Y)$, loss function for a single minibatch (here and below some indices will be omitted) \\ 
$L(\ve\theta,\ve\rho) = \EE_{(X,Y),x} [L(\ve\theta, \ve h(\ldots), X,Y)]$, 
  mean loss over the dataset $(X,Y)$ and $\ve x$, substitution of~\eqref{eq:h-def} is assumed;
  absence of $X,Y$ arguments will assume averaging \\
$L_0(\ve\theta,\ve h,X,Y)$, loss function of the network being pruned \\
$w_l = \sum_c  p_{l,c} / n_l$, estimated fraction of remaining channels at site~$l$ \\
$F(\ve\rho) = F(w_1, \ldots, w_l)$, resource consumption; usually a polynomial over~$\{w_l\}$:
\begin{equation}
  F(\ve\rho) =
    \sum_{l_\text{in}, l_\text{out}} F_{l_\text{in}, l_\text{out}} w_{l_\text{in}} w_{l_\text{out}} + 
    \sum_l G_l w_l
\end{equation}


\subsection{Pruning by learnable channel-wise dropout}
We represent channel-wise pruning as scaling with per-channel factors $\ve h$, which are sampled 
from $[0,1]$ range according to parameters $\ve\rho$~--- see~\eqref{eq:h-def}.
Such scaling is usually inserted at the inputs of convolutional layers, but it can be inserted anywhere where pruning
by zeroing blocks of activations makes sense.

We design $h(\rho, x, \eps)$ function to converge to~\eqref{eq:h-lim} when $\eps\to0$.
Obviously,
\begin{equation}  \label{eq:h-eps0}
  h(\rho, x, \eps=0) = [x<\sigma(\rho)]
\end{equation}
fits for $\eps=0$, with $[\cdot]$ being the indicator function.
This function is discontinuous at $x=\sigma(\rho)$.
We would like $h$ to be Lipschitz-continuous for a fixed $\eps>0$ to use an SGD-based solver,
be continuous and sensible.
To construct such function we shift the separation line $x=\sigma(\rho)$ apart and interpolate linearly in the resulting gap, see Fig.~\ref{fig:rho-x}:
\begin{equation}  \label{eq:h-constr}
  h(\rho,x,\eps) = s\big(x, (1-\eps\kappa)\sigma(\rho-\eps), \eps\kappa + (1-\eps\kappa)\sigma(\rho+\eps) \big)  \text,
\end{equation}
where $s$ is an interpolating function:
\begin{equation}
  s(x,x_0,x_1) = \begin{cases}
    0  &  \text{for $x\leq x_0$,}   \\
    \frac{x-x_0}{x_1-x_0}  &  \text{for $x_0<x<x_1$,}   \\
    1  &  \text{for $x\geq x_1$.}   \\
  \end{cases}
\end{equation}
We set $\eps=0.5$ (in some experiments~$0.25$) and $\kappa=0.04$.
The probability of fractional $h$ is $\approx \eps/2=0.25$ for $\rho=0$, and is $\eps\kappa=0.02$ for $\abs{\rho}\gg1$.
The gap is wider near $\rho=0$ (since $\kappa\ll1$) to allow parameters spend more time in the transitional 
region, while $\rho$ moves from the not-pruned state $\rho\gg1$  towards  the pruned state $(-\rho)\gg1$.
At the same time, the majority (98\%) of channels in $\abs{\rho}\gg 1$ region are completely pruned ($h=0$) or not pruned ($h=1$)~---
to keep the activations in the training mode ($\eps=0.25>0$) close to the activations in the inference mode ($\eps=0$).

We choose a sigmoid function for $p=\sigma(\rho)$ to have plateau regions~--- so that every channel (and its pruning parameter) has 
a burn-in period from the initial value of $\rho=\rhomax\gg 1$ before it starts to get pruned near $\abs{\rho}\sim 1$.
To overcome vanishing gradients problem for~$\rho$, we update $\rho$ with derivatives over~$p$: 
\begin{equation}
  \dd L/\dd p = \big( \dd L/\dd \rho \big) \big/ \big( dp/d\rho \big)  \text.
\end{equation}
To update the pruning parameters $\ve\rho$ with an SGD-based solver, we backpropagate gradients through $h(\rho,x,\eps)$ function.
For easier calculation we approximate $\dd(\EE_x[L_0])/\dd p \approx -\EE_x [\dd L_0/\dd x]$.\footnote{%
This becomes exact for $\eps=0$. Here $\EE_x[\cdot]$ is expectation over random variable~$x$.}  
We use RMSprop-based solver to update the $\ve\rho$ parameters, 
independent of the solver used for the neural network parameters~$\ve\theta$:
\begin{equation}
  D_{l,c}^{i+1} = (1-\delta) D_{l,c}^i  +  \delta \cdot (L'_{0p})^2  \text,
\end{equation}
\begin{equation}
  L'_{p, \text{norm}} = \clip\bigg( \frac{L'_p}{\sqrt{D_{l,c}^{i+1}}}, -3, +3 \bigg)  \text,
\end{equation}
\begin{equation}
  \rho_{l,c}^{i+1} = \clip\big(
    \rho_{l,c}^i - \alpha_\rho L'_{p, \text{norm}},
    -\rhomax, +\rhomax
  \big)  \text,
\end{equation}
where
\begin{equation}
  L'_{0p} = -\sum_n \frac{\dd L_0}{\dd x_{l,c}^{i,n}}  \text,\quad
  L'_p = \sum_n  \bigg(
    -\frac{\dd L_0}{\dd x_{l,c}^{i,n}}
    - \lambdaF \frac{\dd F}{\dd p_{l,c}^{i,n}}
  \bigg)  \text,
\end{equation}
and $\clip(t,a,b)=\max(a, \min(t,b))$ is the clipping function.
We set $\rhomax=12$, $\alpha_\rho=0.03$, and $\delta=1/200$.

\subsection{Updating $\Fsched$}
Expression \eqref{eq:Fsched0} is suggested by the desired invariance of $\Fsched(i)$ 
to the scaling of neural network weights.
Since $\lambdaF$ is a coefficient balancing $L_0$ and $F$, we expect it to scale like $[\text{loss}]/[\text{resources}]$.
For example, if we measure resources in MFLOP instead of GFLOP, $\lambdaF$ is expected to decrease~1000x.
$\lambdaF$ is expected to be independent of weight scale, since $[\theta]$ dimension canceled out.
From the same dimensional consideration we expect loss deterioration due to pruning alone (separately from fine-tuning)
\begin{equation}
  [\frac{\Delta L}{\Delta i}] = \frac{[\text{loss}]}{[\text{resources}]} \cdot \frac{\Delta F}{\Delta i}
\end{equation}
to be approximately constant for schedule~\eqref{eq:Fsched0}.

We update $\Fsched$ using a modified equation~\eqref{eq:Fsched0} to work around the problems
with zero and negative values of~$\lambdaF$:
\begin{equation}
  M^{i} = \begin{cases}
    \frac{\mlr}{\abs{\lambdaF} + 10^{-6}}  &  \text{for $\lambdaF^i<0$,}\\
    +\infty  &  \text{for $\lambdaF\geq0$,}
  \end{cases} 
\end{equation}
\begin{equation}  \label{eq:Fsched}
  \Fsched(i+1) = \Fsched(i) - \clip\Big(\frac{\Fsched^i - F_0}{r}, -M^{i+1}, +M^{i+1}\Big)  \text.
\end{equation}
We set $r=30\,000$ and $F_0=0$.  The term with $r$ limits the initial behavior of $\Fsched$, since initially $\lambdaF$ is close to~$0$.
Parameter $F_0$ is used for the final fine-tuning with constant~$\Fsched$ to improve network quality metric.
$\Fsched(i=0)$ is initialized with $F(\ve\rho^{i=0})$.

\begin{table*}[t]
\begin{center}
\begin{tabular}{ccc}
\hline
$\mu$   &   top-1 accuracy  &  iterations to $500$~MFLOP  \\
\hline\hline
$10^{-4}          $   &  $54.24\%$   &  $ 19\cdot 10^3$  \\
$3.16\times10^{-5}$   &  $54.71\%$   &  $ 52\cdot 10^3$  \\
$10^{-5}          $   &  $55.31\%$   &  $143\cdot 10^3$  \\
$3.16\times10^{-6}$   &  $55.43\%$   &  $422\cdot 10^3$  \\
original net          &  $56.82\%$   &   ---  \\
\hline
\end{tabular}
\end{center}
   \caption{Dependence of top-1 accuracy after pruning to 500 MFLOP on the pruning speed.}
\label{tab:mlr}
\end{table*}
\begin{table*}[t]
\begin{center}
\begin{tabular}{lcccc}
\hline
                    &  GFLOP  &    top-1 accuracy  & top-5 accuracy &  training, epochs  \\
\hline\hline
original AlexNet     & $0.724$  &  $56.8\%$   &  $79.9\%$   &   90  \\
pruned AlexNet, 
  after fine-tuning  & $0.411$  &  $55.8\%$   &  $79.1\%$   &  157  \\
retrained pruned 
configuration of AlexNet  & $0.411$ & $53.1\%$  & $77.1\%$  &  90  \\
\hline
original VGG-16      & $15.47$  &  $68.4\%$   &  $88.4\%$   &       \\
pruned VGG-16, 
  after fine-tuning  &  $3.87$  &  $67.4\%$   &  $88.1\%$   &   75  \\
\hline
\end{tabular}
\end{center}
   \caption{Results of the retraining experiment  and  of the benchmark pruning runs on AlexNet and VGG-16.}
\label{tab:retrain}
\end{table*}
\subsection{Updating $\lambdaF$}
Updating $\lambdaF$ is necessary to keep~\eqref{eq:constr} in balance.
We do this with proportional feedback:
\begin{equation}
  \lambdaF^{i+1} = -\beta\frac{F(\rho^i)-\Fsched(i)}{K}
\end{equation}
with $\beta=0.05$ and $K$ designed to coarsely estimate the change of $F(\rho^{i+1})$ from change in $\lambdaF^{i+1}$:
\begin{equation}
  K = \sum_{l,c} \bigg(\frac{\dd F}{\dd p_{l,c}}\bigg)^2 \cdot \frac{\dd p_{l,c}}{\dd \rho_{l,c}}\cdot \frac{\alpha_\rho}{\sqrt{D_{l,c}}}  \text.
\end{equation}

\subsection{Inference}
For network validation  we  set $\eps=0$~--- this simulates inference with a reduced number of channels,
because channel multipliers $\ve h$ become either $0$, or~$1$.
To get the set of the remaining channels for the chosen pruned snapshot, one can a)~sample $\ve h$ with $\eps=0$, 
and keep the channels with~$h_{l,c}=1$, 
or b)~keep the channels with~$\rho_{l,c}>0$.  
In practice the second approach is better, since it is deterministic.

Let us note, that in practice during pruning the majority of channels are far from~$\rho=0$, 
that is are either completely pruned, or completely not pruned.

\section{Experiments}
A series of experiments was conducted with pruning AlexNet on ImageNet ILSVRC-2012 dataset.
The initial weights were takes from Caffe models~\cite{bib:caffe-models}, initial $F=724.4$~MFLOP
(we only count multiplications in convolutional and FC layers).
The runs were stopped after crossing $F(\ve\rho)=475$~MFLOP.
We didn't fine-tune with fixed $\Fsched$ after this.
The models were tested on the validation set every 2000 iterations (batch size 256),
top-1 accuracy metric was averaged over the range $F \in (480,520)$~MFLOP on a FLOPs-accuracy plot.

We split each grouped convolutional layer in AlexNet into two ungrouped convolutional layers,
inserting channel concatenations and slicings in proper places.

\subsection{Repeatability}
To measure the accuracy of this metric, we did 7 pruning runs with the same settings, 
and used $\sigma_{n-1}$ estimator, 
which resulted in standard deviation of $\sigma_{\text{top-1}} = 0.03\%$.

\subsection{Dependence on pruning speed}
We conducted 4 pruning runs with different values of $\mlr$ summarized in table~\ref{tab:mlr}.
This shows that from $\mlr=10^{-4}$ to $\mlr=10^{-5}$ accuracy consistently improves,
and from $\mlr=10^{-5}$ to $3.16\times10^{-6}$ it saturates.

As a limiting case of quick pruning we consider training the pruned configuration of AlexNet
starting with random initialized weights.
We take channel configuration of our best pruned AlexNet@ILSVRC-2018 model obtained by pruning and fine-tuning for 157 epochs 
as a ``template'', and train it following the same procedure as for the original Caffe AlexNet.
The results are summarized in table~\ref{tab:retrain}.

These results are compatible with the intuition that knowledge transfer from larger to smaller models improves with more gradual profiles of~$F(i)$.

\begin{figure*}[t]
\begin{center}
\includegraphics{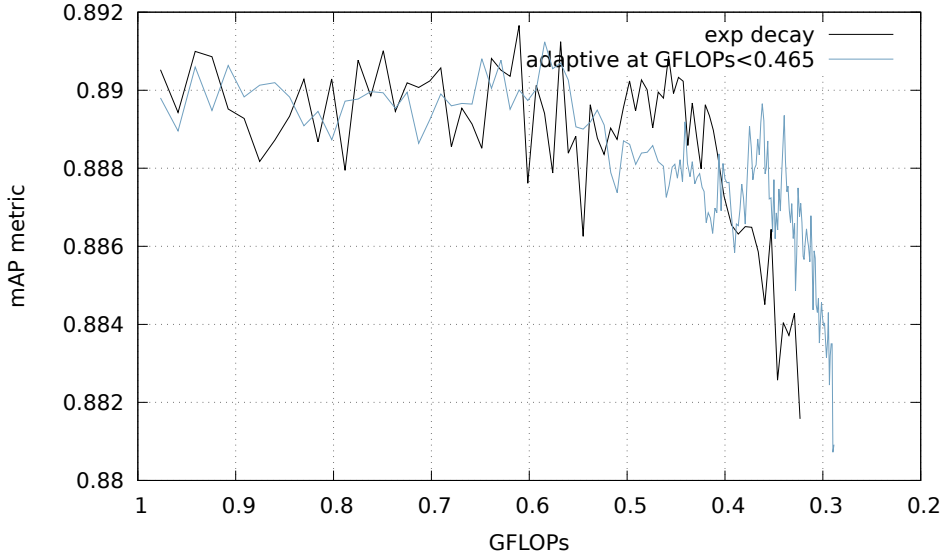}
\end{center}
   \caption{FLOPs-accuracy curves for A) the exponential FLOPs schedule; B) including the adaptive FLOPs schedule. 
   Both curves start with exponential decay ($F_0=0.1$~GFLOP), curve~B switches to the adaptive 
   schedule ($\mu=5\times 10^{-6}$) at $F=0.465$~GFLOP.}
\label{fig:knee}
\end{figure*}
\subsection{Dependence on the shape of schedule~$\Fsched(i)$}
Here we compare FLOPs-quality pruning curves obtained by A) the exponential relaxation of $\Fsched$ versus
B) a schedule defined by~\eqref{eq:Fsched} in the region $F<0.465$~GFLOP. 
Curve B before the switch was obtained with the same settings as curve A.\footnote{%
The reason for this switch is that to get the practical result in a limited time we quickly pruned with exponential decay to the point before
quality starts to drop, and switched to the adaptive mode.}
We didn't rerun this experiment on AlexNet, it was conducted on a custom SSD-based~\cite{bib:ssd} object detector.

Comparison of the pruning curves on Figure~\ref{fig:knee} shows that for the faster pruning schedule~A 
quality starts to fall quickly earlier that for the slower schedule~B: curve~A changes slope near $F=0.41$~GFLOP, 
curve~B near $F=0.33$~GFLOP.
In the region between these two points, the approximately constant pruning speed ($\Delta \Fsched(i)/\Delta i$) of curve A becomes
higher than the pruning speed required to maintain network quality,
while the slower speed of curve B is still below that threshold.
This means that the pruning speed required to maintain network quality changes through the pruning process

We interpret the sharp change in slope of these curves as a change in knowledge transfer process from the saturated mode (pruning speed is low enough),
to a highly unsaturated mode (pruning speed is too high), in the following sense:
too high/low enough to replace the roles of the pruned out channels by fine-tuning the remaining channels.

\begin{figure*}[t]
\begin{center}
\includegraphics{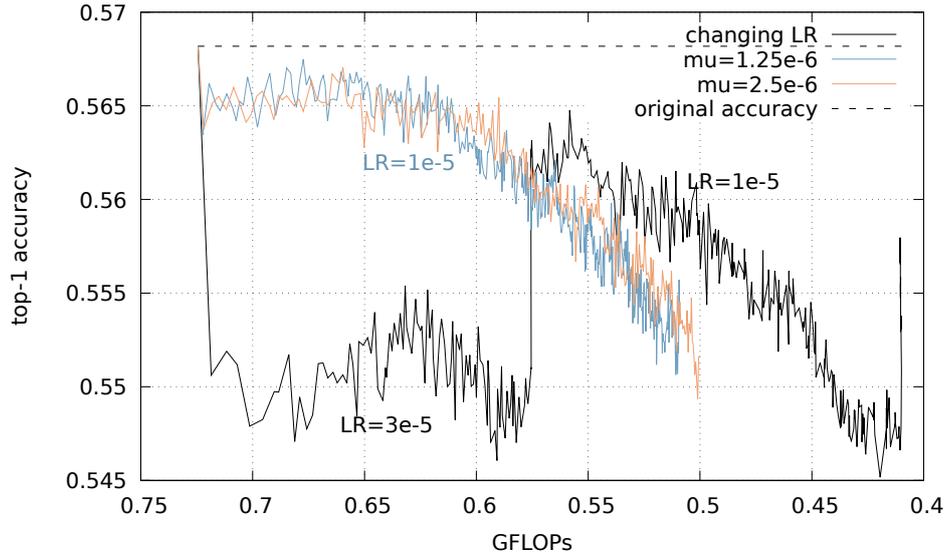}
\end{center}
   \caption{FLOPs-accuracy curves for a) the best AlexNet pruning run with non-constant weight learning rate; b) with constant learning rate.}
\label{fig:alexnet}
\end{figure*}
\subsection{Pruning AlexNet @ ILSVRC-2012}
We did a manual hyperparameter search by restarting pruning from a snapshot to get faster feedback.
As a result our best pruned model (by FLOPs at fixed accuracy deterioration) 
turned out to contain sections with different values of weight learning rate: $\text{LR}=3\times10^{-6}$ (ADAM solver) up to $0.576$~GFLOP, 
then $\text{LR}=10^{-5}$ with $\mu=5\times10^{-6}$.
Interestingly, we couldn't reach the same accuracy using constant weight learning rate ($\text{LR}=10^{-5}$), 
even with much slower pruning (Figure~\ref{fig:alexnet}).

Another unexpected point is that the weight learning rate that resulted in the best final accuracy provided 
a worse accuracy metric at the initial stages of pruning, 
i.e. the better ``state of pruning'' was not reflected in accuracy metric.
This suggests that there is a room for improvement by changing weight learning rate through pruning.

For the summary of the best pruning run (with the additional fine-tuning), see Table~\ref{tab:retrain}.

\subsection{Pruning VGG-16 @ ILSVRC-2012}
We did a single pruning run with VGG-16 with guessed parameters (with the additional fine-tuning), results in table~\ref{tab:retrain}.
The weights trained by the authors of VGG-16~\cite{bib:vgg-16} were used for the initialization.
Our result is close to the best published result of channel-wise pruning of VGG-16 that we known in~\cite{bib:HWang2017SPP},
and is better than~\cite{bib:Molchanov2016} 
(judging by their Figure~9 and taking into account the the 2x difference in the definition of FLOPs).

\section{Conclusion}
We suggest the first method for channel-wise pruning of neural networks
that combines:
\begin{enumerate}
\item
Constrained optimization by the method of Lagrange multipliers
to control and limit resource (e.g. FLOPs) consumption by the pruned network~---
this enables pruning by a simple addition of the Lagrangian term to the loss function 
and insertion of pruning layers at the desired pruning sites of the network.
This method doesn't limit the neural network to some specific task or kind of loss function.
\item
Good separation between the pruning parameters and the weights of the neural network, as well as between the corresponding solvers.
This improves modularity: the weight solver can be controlled independently from the pruning solver.
\item
Explicit and adaptive control over the schedule of the amount of available resource (FLOPs) vs. pruning iteration.
We show that this schedule affect quality of the pruned network.
\end{enumerate}

Our method can potentially constrain any resource that depends smoothly on the numbers of remaining channels at the pruning sites,
like the number of FLOPs, size of weights, size of activations, and their combinations.

The method can be (and was) applied to branched networks (GoogleNet+SSD), depthwise convolutions (MobileNet~V1+SSD),
object detection (SSD, Faster-RCNN) and classification (AlexNet, VGG-16) networks.
The method can be easily extended to support coordinated pruning of the corresponding channels in the layers neighbouring skip-connections (ResNets) 
by sharing parameters between the pruning layers.



{\small
\bibliographystyle{ieee}
\bibliography{pruning_bib}
}

\end{document}